\documentclass[runningheads]{llncs}

\usepackage{cv4dc}

\usepackage{cv4dcabbrv}

\usepackage{graphicx}
\usepackage{booktabs}

\usepackage[accsupp]{axessibility}  

\usepackage[pagebackref,breaklinks,colorlinks,citecolor=cvdcblue]{hyperref}

\usepackage{orcidlink}
\usepackage{cite}
\usepackage{amsmath,amssymb,amsfonts}
\usepackage{algorithmic}
\usepackage{graphicx}
\usepackage{textcomp}
\usepackage{xcolor}
\usepackage{subcaption}
\usepackage{nameref}
\usepackage{caption}
\usepackage{hyperref}
\usepackage{array}
\usepackage{multirow}
\usepackage[utf8]{inputenc}
\usepackage{tabularx}
\usepackage{url}
\begin{document}

\title{IncepDeHazeGAN: Novel Satellite Image Dehazing} 


\author{Tejeswar Pokuri,
Shivarth Rai}



\maketitle

\begin{abstract}
Dehazing is a technique in computer vision for enhancing the visual quality of images captured in cloudy or foggy conditions. Dehazing helps to recover clear, high-quality images from haze-affected remote sensing data. In this study, we introduce IncepDeHazeGAN, a novel  Generative Adversarial Network (GAN) involving Inception block and multi-layer feature fusion for the task of single-image dehazing. Utilizing the Inception block allows for multi-scale feature extraction. On the other hand, the multi-layer feature fusion design achieves efficient reuse of features as the features extracted at different convolution layers are fused several times. Grad-CAM XAI technique has been applied to our network, highlighting the regions focused on by the network for dehazing and its adaptation to different haze conditions. Experiments demonstrate that our network achieves state-of-the-art results in several datasets. 
\end{abstract}

\section{Introduction}
Haze caused by atmospheric conditions that scatter light lead to loss in detail, reduced contrast and color distortion in images. This significantly affects applications where high-quality images are crucial for decision making. Dehazing is a field in computer vision and image processing that aims to extract high-quality scenes from hazy photos [1]. Dehazing has applications in autonomous driving, surveillance systems, image enhancement and, remote sensing and satellite imagery [2]. This study focuses on the application of dehazing techniques on satellite imagery. 

The atmospheric scattering model [3, 4] is a widely used physics based framework for developing algorithms for image dehazing. The model assumes the atmosphere has a homogenous distribution of particles with isotropic properties. The model is expressed by the formula:
\newline
\begin{equation}
    I(x) = J(x)t(x) + A(1 - t(x))
\end{equation}
\newline
where \(I(x)\) is an observed hazy image \(J(x)\) is the haze-free image (“clean image”) to be recovered. There are two critical parameters: \(A\) denotes the global atmospheric lighting, and \(t(x)\) is the transmission matrix defined as:
\newline
\begin{equation}
    t(x) = e^{-\beta d(x)}
\end{equation}
\newline
where $\beta$ is the atmospheric scattering coefficient, and \(d(x)\) is the distance between the object and the camera. The goal of dehazing algorithms based on the atmospheric scattering model is to estimate \(t(x)\) and \(A\), and then use these to recover the haze-free image \(J(x)\).

Prior-based methods of image dehazing, such as [5], introduce diverse assumptions to help address the challenges of the ill-posed atmospheric scattering model. [6] suggests two new priors: the local patchwise minimal values prior (MinVP) and the local patchwise maximal values prior (MaxVP). It has been shown that these priors are more effective in producing transmission maps that are comparable to those obtained by DCP and BCP. Trung et al. [7] propose Color Ellipsoid Prior (CEP), a dehazing technique based on the generation of colour ellipsoids in RGB space. The two main elements of this method are the incorporation of fuzzy segmentation and the statistical fitting of hazy pixel clusters into ellipsoids. 

In recent years, several learning-based methods such as CNNs, GANs and ViT have been used for the task of single image dehazing. Many studies have successfully applied Generative Adversarial Networks (GAN) [8] to image dehazing and achieved excellent results. The application of GANs in dehazing leverages their ability to generate high-quality images by learning the underlying distribution of clear images from hazy inputs. GANs consist of two neural networks: the generator and the discriminator. The generator learns to produce new data indistinguishable from ground truth by mapping input noise to data space. The discriminator’s role is to distinguish between fake data produced by the generator and the real data. It is trained to maximise its ability to correctly classify real and fake data. This adversarial training continues until the generator is able to produce data which the discriminator cannot reliably classify.

In this work, we introduce a simple yet effective image dehazing network, IncepDeHazeGAN. To address the challenges in efficiently combining feature data from various levels, our model utilises Inception module [9]  and Residual connections [10]. The Inception module applies multiple convolutional filters with different kernel sizes (e.g., 1x1, 1x3, 3x3 in our network) simultaneously to the input data. This enables multi-level feature extraction where the network captures local and global features at several levels. Residual connections allow feature maps to bypass one or more intermediate layers and connect directly to a subsequent layer. They help deal with vanishing and exploding gradients, and their regularising property helps avoid model collapse in GANs [11]. IncepDeHazeGAN uses a multi-level feature fusion design, which helps in mitigating information loss due to downsampling. Fusing feature maps from different layers enhances feature representation as low-level and high-level features are learned simultaneously [12]. We summarize our contributions as follows:

\begin{itemize}
    \item We propose a novel architecture that shifts focus from attention-blocks onto simple multi-layer feature extraction and skip connections for the task of image dehazing. This allows our network to be lightweight and achieve state-of-art results with a GAN-based architecture.
    \item Utilize Explainable AI technique, Grad-CAM, for the purpose of analyzing and verifying the decision making process of the network.
\end{itemize}

\section{Related Works}
Image dehazing is an important research topic in computer vision. With the rapid advancement of learning-based technology, several deep-learning methods have been applied to this task. He et al.[13] introduce LRSDN, a lightweight CNN- based dehazing network with less than 0.1M parameters. This proposed architecture consists of two main components, the Axial Depthwise Convolution and Residual Learning Block (ADRB) and the Hybrid Attention Block (HAB). ADRB utilizes axial depthwise convolutions and residual connections to enhance feature extraction and mitigate the gradient vanishing problem. HAB integrates channel attention and pixel attention mechanisms. Channel attention enhances relevant features while suppressing less important ones. Pixel attention incorporates observational priors allowing for better reconstruction of areas covered in deep haze. The lightweight design of this architecture makes it suitable for real-time applications. Du et al.[14] propose AU-Net, an end-to-end asymmetric U-Net dehazing network. This network incorporates optimization of physical parameters with deep-learning techniques to enhance image clarity. First, Unet is used to estimate key physical parameters, including atmospheric light \textit{A} and transmittance map \textit{T}. Next, a rough dehazing image \textit{J} is deduced using \textit{A} and \textit{T}. Finally, AU-Net utilizes self-attention with depth (SAD) and channel attention (CA) modules to refine \textit{J} once again. SAD focuses on merging deep semantic information with shallow detail features, while CA emphasizes the importance of different channels in the feature maps. Wen et al.[15] propose RSHazeNet, an encoder-minimal and decoder-minimal network, where the encoding and decoding stages are down-sampling and up-sampling operations. Skip connections are utilized to maintain information flow. For the task of merging features at the same level, the authors develop the intra-level transposed fusion module (ITFM). ITFM employs adaptive pooling and 1x1 convolution to generate attention maps, which are a fusion of features from different sources and capture contextual dependencies. The authors also introduce cross-level multiscale interaction module (CMIM) and multi-view progressive extraction block (MPEB). CMIM uses transposed self-attention to compute attention maps, allowing features from different resolutions to interact. This addresses the information loss due to repeated sampling operations. MPEB partitions the input features into four components, each processed by convolutions of varying kernel sizes and dilation factors. MPEB enables the network to learn multi-view features, capturing diverse spatial information. The All-in-One Dehazing Network (AOD-Net) introduced by Boyi et al.[16] utilizes a re-formulated atmospheric scattering model. The K-estimation module consists of five convolutional layers that estimate a variable \(K(x)\). This variable integrates the effects of transmission matrix and atmospheric light. Following K-estimation, element-wise operations compute the final dehazed image using the estimated \(K(x)\). The architecture captures multi-scale features improving reconstruction of fine details in the output images. In [17], Yufeng et al. introduce FCTF-Net (First-Coarse-Then-Fine Network). FCTF-Net employs a two-stage dehazing process. The Coarse-Scale dehazing process utilizes a encoder-decoder structure to extract multi-scale features and produce initial results, This stage incorporates Residual Dense Blocks to enhance information flow and facilitate feature reuse. The Fine-Scale dehazing stage utilizes information from the coarse stage to refine the initial output and improve dehazing performance. FCTF-Net architecture includes Channel Attention (CA) mechanism to adjust importance of different feature channels based on their relevance to haze distribution. This mechanism allows the model to focus on varying haze characteristics at different scales, improving overall dehazing ability.

\section{Datasets}
In this study, we utilized two publicly available datasets namely Haze1k and RICE dataset to train and evaluate our proposed model for Satellite Image Dehazing.
\subsection{Haze1k}
The Haze1k dataset [22] is a widely recognized resource for remote sensing image dehazing. The dataset consists of synthetic hazy images with corresponding clear images, and SAR images. The original clear images are sourced from the GF-2 satellite, while the SAR images come from the GF-3 satellite. In line with standard practices for optical remote sensing image dehazing, only the RGB hazy and clear images were used, excluding SAR images as additional input. All images are at a resolution of 512 × 512 pixels. The dataset is split into four sections: train, test thin, test moderate, and test thick, with 900 images allocated for training and 45 images for each testing category of cloud thickness.
\begin{figure}[!ht]
    \centering
    \begin{subfigure}[b]{0.4\linewidth}
        \centering
        \includegraphics[width=\linewidth]{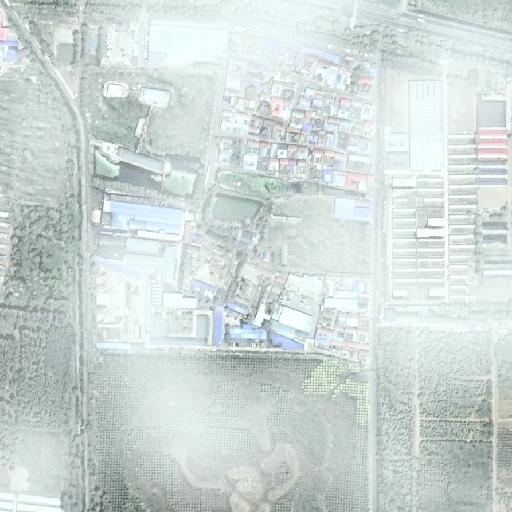}
        \caption{Hazy Image}
        \label{fig:image1}
    \end{subfigure}
    \hspace{0.05\linewidth} 
    \begin{subfigure}[b]{0.4\linewidth}
        \centering
        \includegraphics[width=\linewidth]{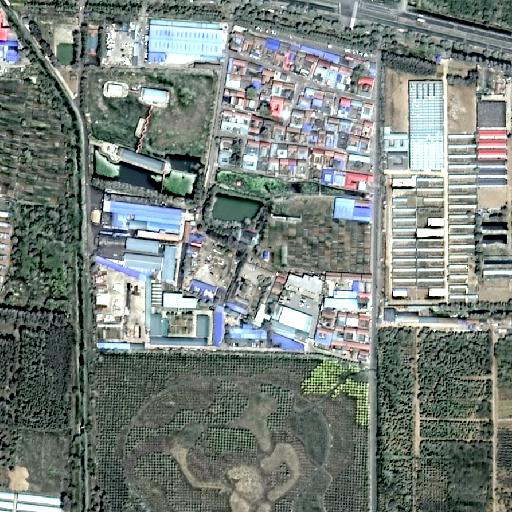}
        \caption{Clear Image}
        \label{fig:image2}
    \end{subfigure}
    \caption{Sample Pair from Haze1k dataset}
    \label{fig:side_by_side}
\end{figure}
\subsection{RICE}
The RICE (Remote sensing Image Cloud rEmoving)[23] dataset is divided into two parts: RICE-1 and RICE-2. For our study, we are using RICE-1. This subset includes 500 pairs of images, each consisting of a cloud-covered and a cloudless image, both sourced from Google Earth by adjusting the cloud layer settings. All images are 512x512 pixels in size. We allocated 90\% of the data for training and 10\% for testing.
\begin{figure}[!ht]
    \centering
    \begin{subfigure}[b]{0.4\linewidth}
        \centering
        \includegraphics[width=\linewidth]{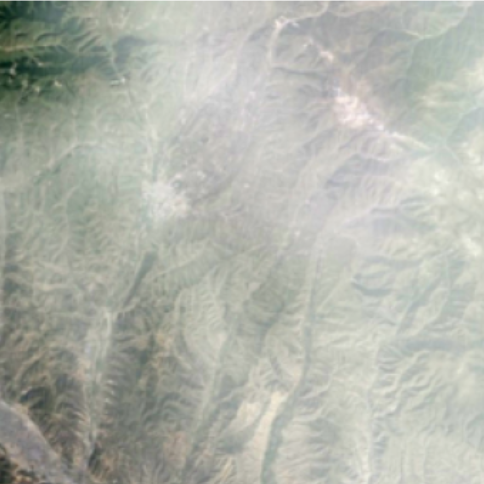}
        \caption{Hazy Image}
        \label{fig:image3}
    \end{subfigure}
    \hspace{0.05\linewidth} 
    \begin{subfigure}[b]{0.4\linewidth}
        \centering
        \includegraphics[width=\linewidth]{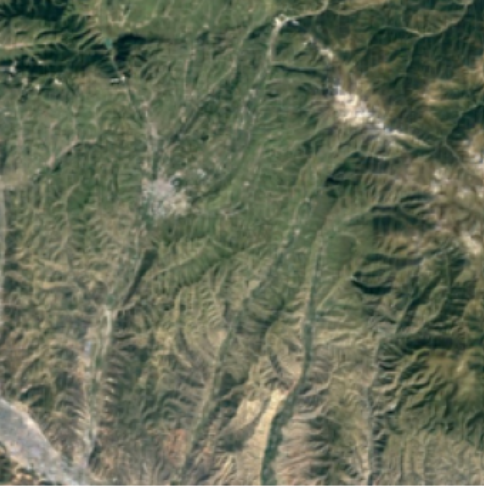}
        \caption{Clear Image}
        \label{fig:image4}
    \end{subfigure}
    \caption{Sample Pair from RICE dataset}
    \label{fig:side_by_sid}
\end{figure}
\section{The Methodology: IncepDeHazeGAN}
\begin{figure}[!ht]
    \centering
    \includegraphics[width=\textwidth]{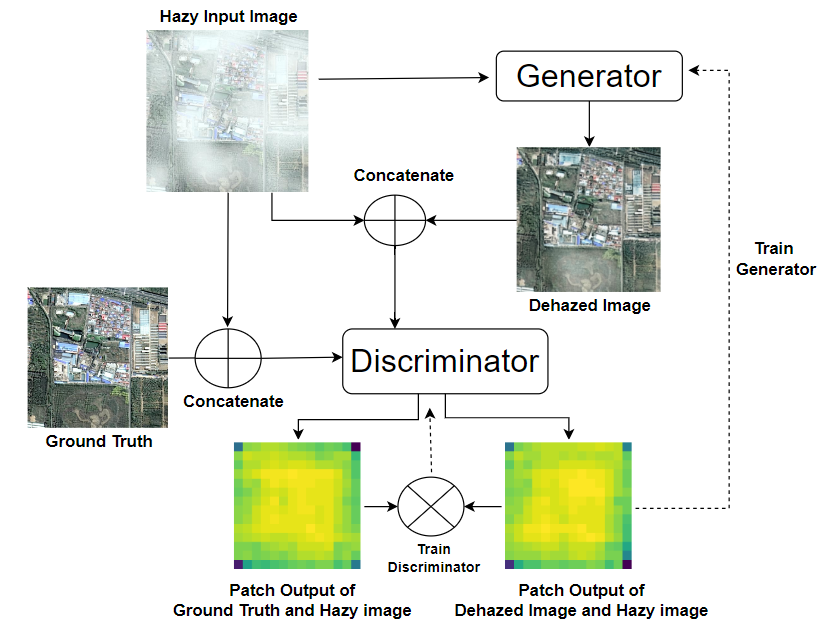}
    \caption{Model overview}
    \label{fig:model_overview}
\end{figure}

Our proposed Model follows a GAN (Generative Adversial Networks) based approach to image dehazing. The network diagram is \cref{fig:model_overview}
\subsection{Generator}
The Generator is designed as a encoder-decoder structure with incorporating Inception Blocks [12], for enchanced feature extraction. Generator takes an hazy image as input and give corresponding dehazed image as output.
The proposed generator network architecture is divided into three parts namely: (1) Encoder block
(2) Inception Module, and (3) Decoder block.
Encoder/Decoder block consists of simple convolution/deconvolution layer followed by nonlinear activation function (ReLU). The Inception Blocks uses different kernel sized convolutions (1x1, 1x3, 3x1, 3x3) to extract features.
We have used the residual connections, as used in [14]. The major purpose behind employing skip connections in the architecture was to share the low-level features learned at initial convolution layers to the deconvolution layers and reduce information loss due to downsampling operations. This concatenation of feature maps helped
them to generate the prominent edge information in the output image. Inspired by this, we used multi-layer feature fusion to obtain better results.
\begin{figure}[!ht]
    \centering
    \includegraphics[width=\columnwidth]{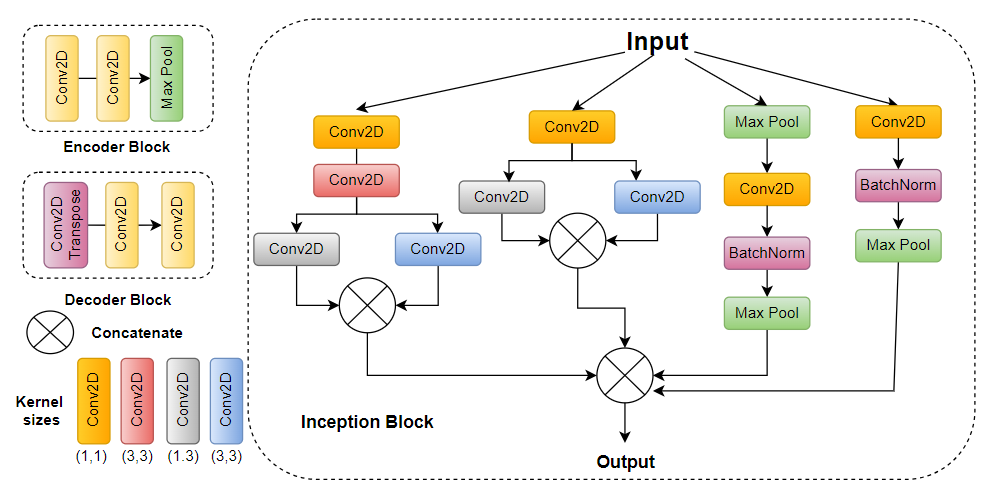}
    \includegraphics[width=\columnwidth]{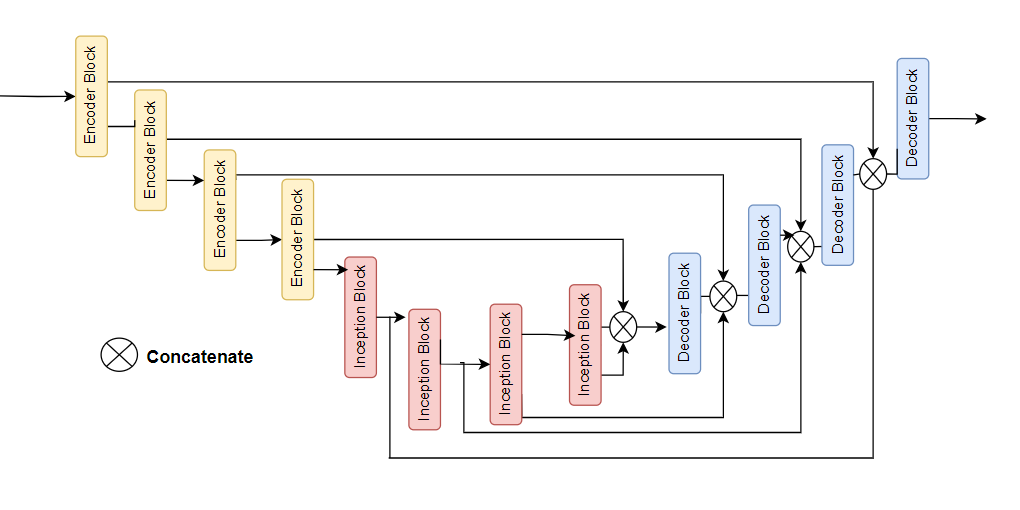}
    \label{fig:gen}
    \caption{IncepDehazeGan Generator}
\end{figure}
In the training of the generator, we employ a combination of adversarial loss and mean absolute error. The adversarial loss, or GAN loss, is computed based on the output of the discriminator and the target labels of ones, encouraging the generator to produce realistic outputs that can deceive the discriminator. This loss is defined as:
\newline
\begin{equation}
\mathcal{L}_{\text{GAN}}(x, y) = {\textstyle -\frac{1}{N} \sum_{i=1}^{N} \left[ y_i \log\left(\sigma(x_i)\right) + (1 - y_i) \log\left(1 - \sigma(x_i)\right) \right]}
\end{equation}
\newline
In addition to the adversarial loss, we incorporate a mean absolute error (L1 loss) between the generated image and the target image. This loss term helps to maintain the fidelity of the generated output relative to the ground truth and is defined as:
\newline
\begin{equation}
\mathcal{L}_{\text{L1}} = \frac{1}{N} \sum_{i=1}^{N} | y_i - \hat{y}_i |
\end{equation}
\newline
The total generator loss is a weighted sum of the adversarial loss and the L1 loss, where \(\lambda\) is a hyperparameter controlling the trade-off between these two losses. In our implementation, \(\lambda\) is set to 100:
\newline
\begin{equation}
\mathcal{L}_{\text{gen}} = \mathcal{L}_{\text{GAN}} + \lambda \cdot \mathcal{L}_{\text{L1}}
\end{equation}
\newline
\subsection{Discriminator}
The discriminator consists of six convolutional blocks, each containing a convolutional layer with leaky ReLU activation and a batch normalization layer. It produces a single-channel matrix with sigmoid activation.

Inspired by [24], we concatenate the input image with either the ground truth or the generator's output before passing it to the discriminator. This approach allows the discriminator to effectively evaluate the plausibility of the generated image.
\begin{figure}[!ht]
    \centering
    \includegraphics[width=\columnwidth]{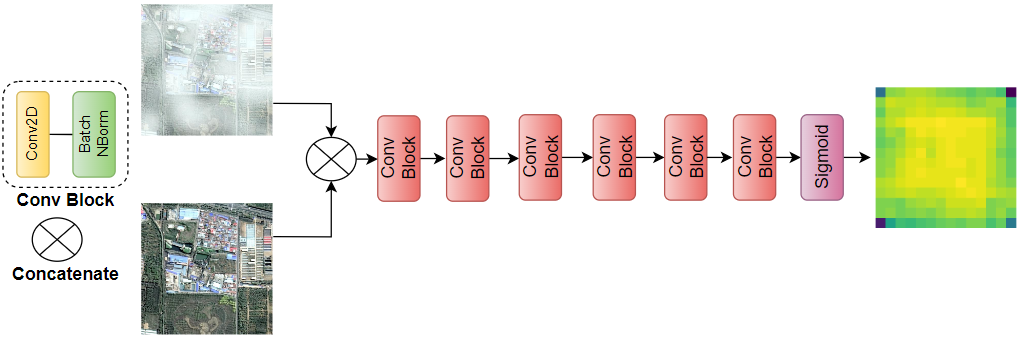}
    \label{fig:di}
    \caption{IncepDehazeGan Discriminator}
\end{figure}
Furthermore, our discriminator outputs a matrix instead of a single value, which encourages the generator to produce realistic local features and textures across the image. By focusing on small regions for each pixel, the discriminator also helps mitigate overfitting.

For training the Discriminator, we use Binary Cross Entropy to distinguish between Ground Truth and Output produced by generator. BCE loss has 2 components - real loss and generated loss.

Real loss is calculated by comparing discriminator's output of ground truth images with ones, indicating that the discriminator should correctly identify these images as real.
\newline
\begin{equation}
\begin{aligned}
\mathcal{L}_{\text{real}} &= -\frac{1}{N} \sum_{i=1}^{N} \log\left(\sigma(D(x_i))\right)
\end{aligned}
\end{equation}
\newline
Generated loss is calculated by comparing discriminator's output of Generated images with zeros, indicating that the discriminator should correctly identify these images as fake.
\newline
\begin{equation}
\begin{aligned}
\mathcal{L}_{\text{fake}} &= -\frac{1}{N} \sum_{i=1}^{N} \log\left(1 - \sigma(D(G(z_i)))\right)
\end{aligned}
\end{equation}
\newline
The total discriminator loss is defined as sum of t real loss and generated loss
\newline
\begin{equation}
\begin{aligned}
\mathcal{L}_{\text{D}} &= \mathcal{L}_{\text{real}} + \mathcal{L}_{\text{fake}}
\end{aligned}
\end{equation}
\subsection{Explainable AI}
Explainable AI(XAI) refers to methods developed to interpret output of deep learning models. XAI aims to address this by providing insights into how models make decisions, thereby increasing trust, transparency and accountability. In this work, XAI has been applied to our image dehazing network, helping understand where the model is focusing on the input image for recovering the dehazed output.

Gradient-weighted Class Activation Mapping (Grad-CAM)[21] is an XAI technique used to visualize and interpret results, particularly for image-based tasks. Grad-CAM uses the gradients of any target flowing into the final convolutional layer to produce a coarse localization map highlighting the important regions in the image for predicting the concept. From a high-level, we take an image as input and select a layer in the model architecture where we want grad-cam output. We then run the input through the model, take the layer output and loss value. Next, we find the gradient of the output of our desired model layer w.r.t. the model loss. From there, we take sections of the gradient which contribute to the prediction. This is followed, reducing, resizing, and re-scaling the heat map so that it can be overlaid on the original image.

In the figure below, the Grad-CAM image overlaid on the input image clearly shows that the model is primarily focusing on the hazy areas, indicating that it is effectively performing the intended task.
\begin{figure}[!ht]
    \centering
    \includegraphics[width=\textwidth]{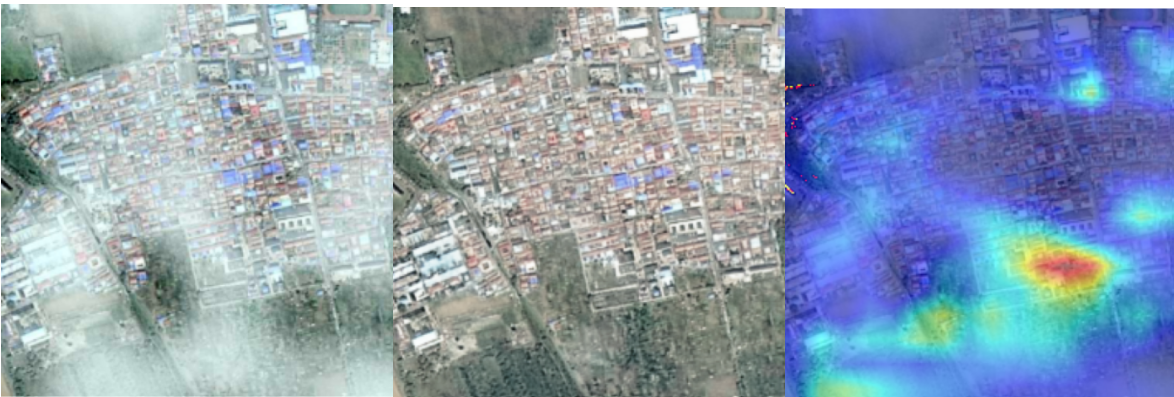}
    \includegraphics[width=\textwidth]{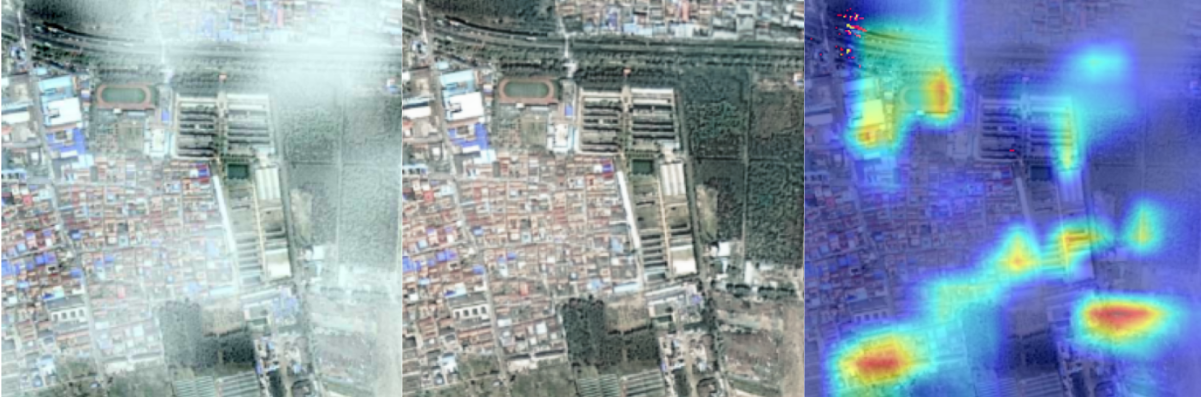}
    \label{fig:try}
    (a)Hazy Image
    \hspace{18mm} 
    (b)DeHazed Image
    \hspace{8mm} 
    (c)GRAD-CAM output
    \caption{An example visualization from GradCAM highlights that the model predominantly concentrates on the hazy regions.}
\end{figure}
\section{Experimental Study}
We utilized two publicly available datasets to evaluate our model, where it outperforms several other models both quantitatively and qualitatively.
\begin{figure*}[ht]
    \centering
    \begin{subfigure}[b]{0.065\textwidth}
        \includegraphics[width=\textwidth]{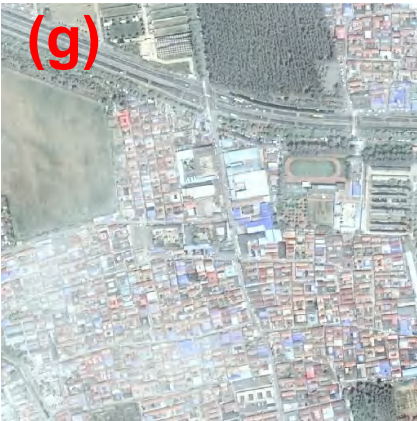}
    \end{subfigure}
    \begin{subfigure}[b]{0.065\textwidth}
        \includegraphics[width=\textwidth]{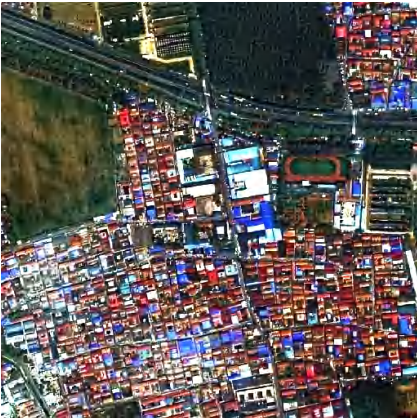}
    \end{subfigure}
    \begin{subfigure}[b]{0.065\textwidth}
        \includegraphics[width=\textwidth]{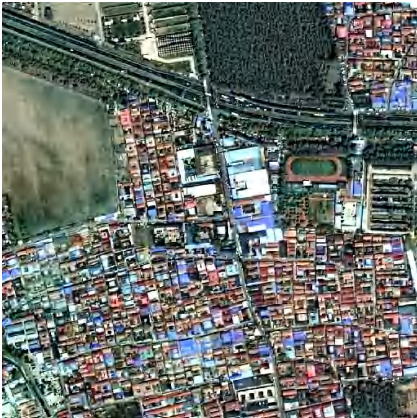}
    \end{subfigure}
    \begin{subfigure}[b]{0.065\textwidth}
        \includegraphics[width=\textwidth]{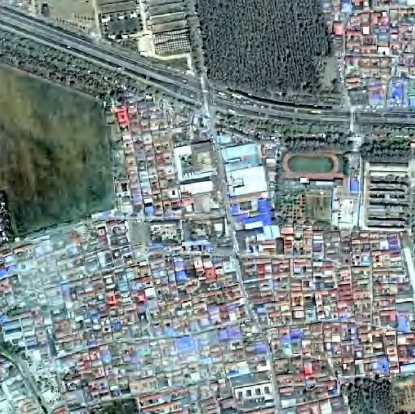}
    \end{subfigure}
    \begin{subfigure}[b]{0.065\textwidth}
        \includegraphics[width=\textwidth]{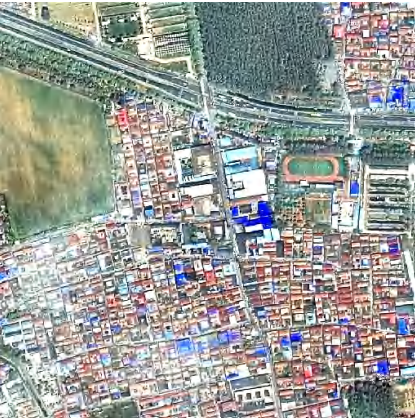}
    \end{subfigure}
    \begin{subfigure}[b]{0.065\textwidth}
        \includegraphics[width=\textwidth]{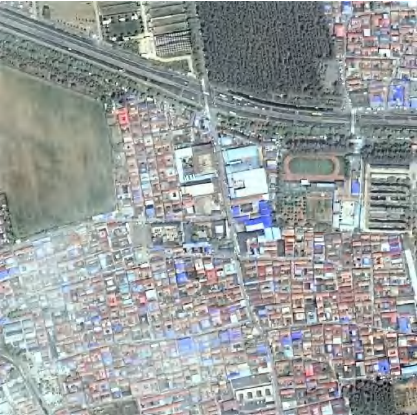}
    \end{subfigure}
    \begin{subfigure}[b]{0.065\textwidth}
        \includegraphics[width=\textwidth]{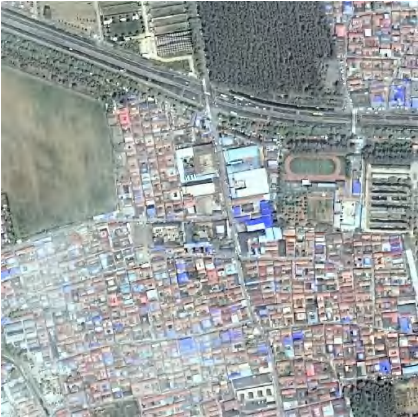}
    \end{subfigure}
    \begin{subfigure}[b]{0.065\textwidth}
        \includegraphics[width=\textwidth]{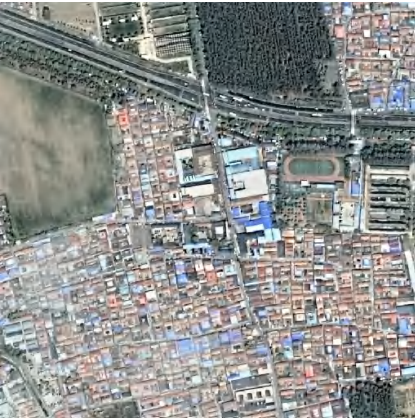}
    \end{subfigure}
    \begin{subfigure}[b]{0.065\textwidth}
        \includegraphics[width=\textwidth]{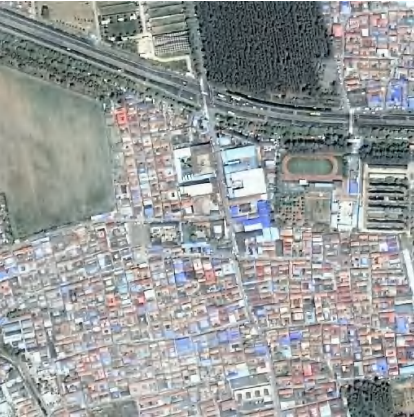}
    \end{subfigure}
    \begin{subfigure}[b]{0.065\textwidth}
        \includegraphics[width=\textwidth]{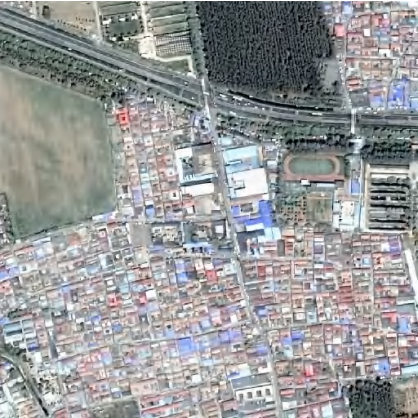}
    \end{subfigure}
    \begin{subfigure}[b]{0.065\textwidth}
        \includegraphics[width=\textwidth]{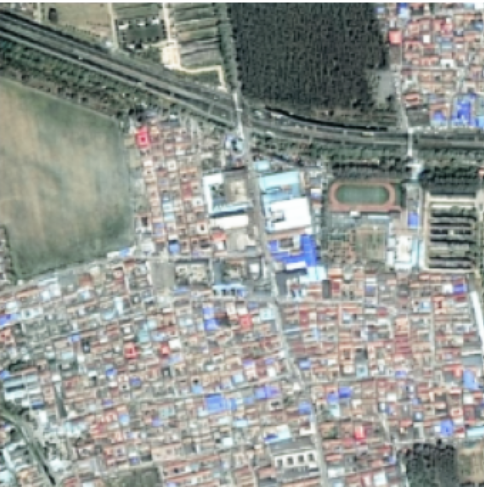}
    \end{subfigure}
    \begin{subfigure}[b]{0.065\textwidth}
        \includegraphics[width=\textwidth]{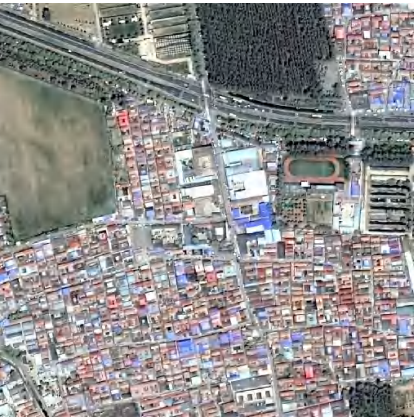}
    \end{subfigure}
    
    \vspace{0.1cm}
    
    \begin{subfigure}[b]{0.065\textwidth}
        \includegraphics[width=\textwidth]{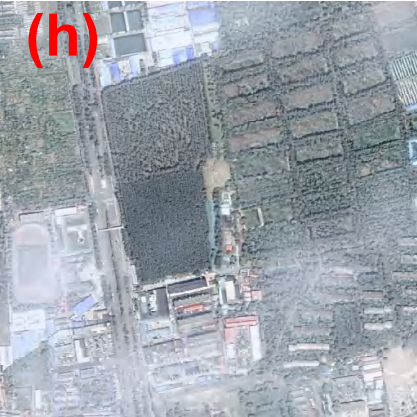}
    \end{subfigure}
    \begin{subfigure}[b]{0.065\textwidth}
        \includegraphics[width=\textwidth]{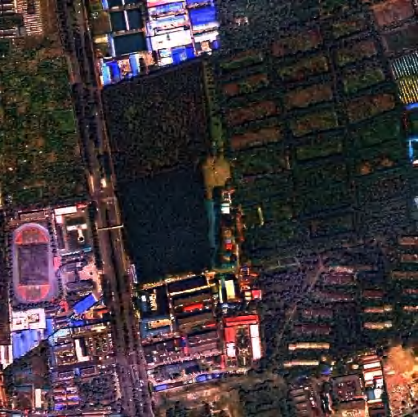}
    \end{subfigure}
    \begin{subfigure}[b]{0.065\textwidth}
        \includegraphics[width=\textwidth]{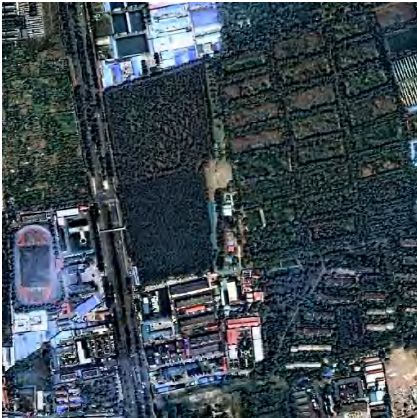}
    \end{subfigure}
    \begin{subfigure}[b]{0.065\textwidth}
        \includegraphics[width=\textwidth]{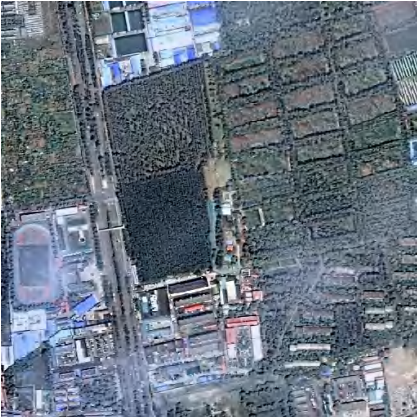}
    \end{subfigure}
    \begin{subfigure}[b]{0.065\textwidth}
        \includegraphics[width=\textwidth]{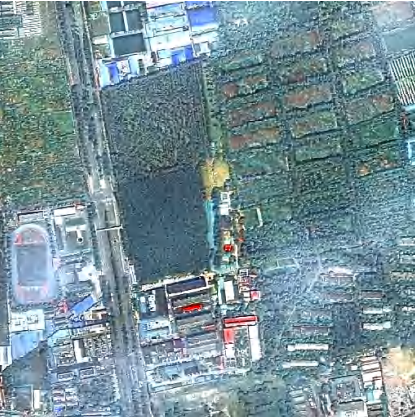}
    \end{subfigure}
    \begin{subfigure}[b]{0.065\textwidth}
        \includegraphics[width=\textwidth]{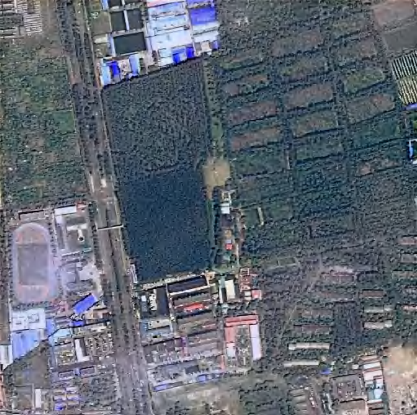}
    \end{subfigure}
    \begin{subfigure}[b]{0.065\textwidth}
        \includegraphics[width=\textwidth]{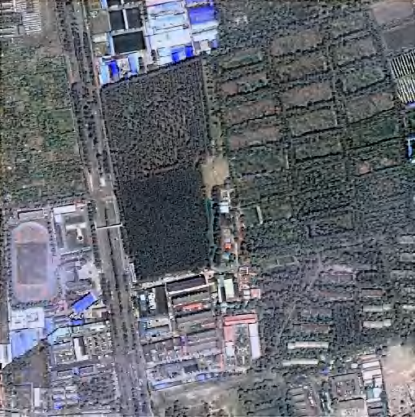}
    \end{subfigure}
    \begin{subfigure}[b]{0.065\textwidth}
        \includegraphics[width=\textwidth]{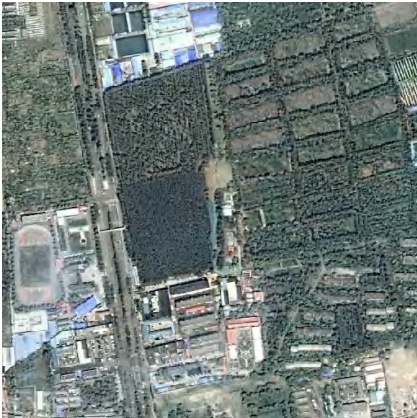}
    \end{subfigure}
    \begin{subfigure}[b]{0.065\textwidth}
        \includegraphics[width=\textwidth]{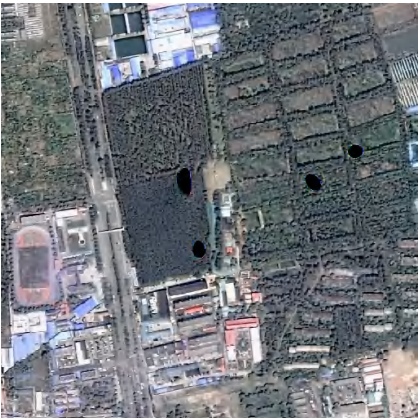}
    \end{subfigure}
    \begin{subfigure}[b]{0.065\textwidth}
        \includegraphics[width=\textwidth]{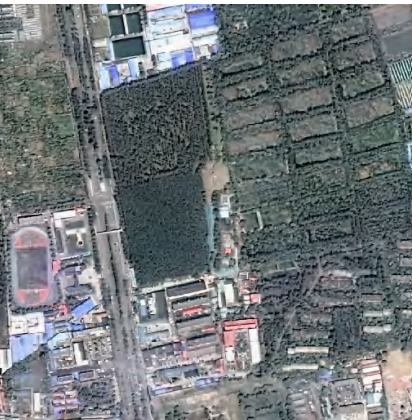}
    \end{subfigure}
    \begin{subfigure}[b]{0.065\textwidth}
        \includegraphics[width=\textwidth]{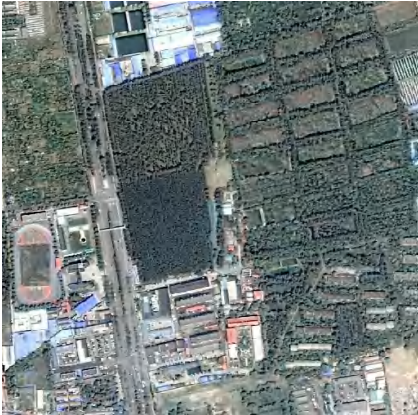}
    \end{subfigure}
    \begin{subfigure}[b]{0.065\textwidth}
        \includegraphics[width=\textwidth]{Haze1k-mod/GT.png}
    \end{subfigure}
    
    \vspace{0.1cm}
    
    \begin{subfigure}[b]{0.065\textwidth}
        \includegraphics[width=\textwidth]{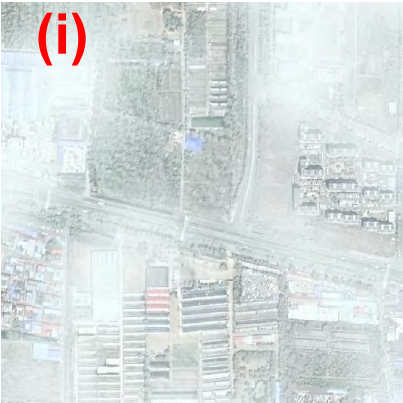}
        \caption*{Input}
    \end{subfigure}
    \begin{subfigure}[b]{0.065\textwidth}
        \includegraphics[width=\textwidth]{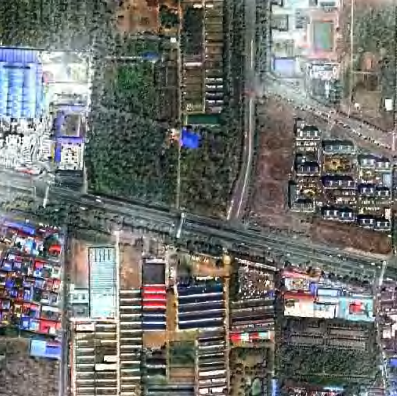}
        \caption*{CEP}
    \end{subfigure}
    \begin{subfigure}[b]{0.065\textwidth}
        \includegraphics[width=\textwidth]{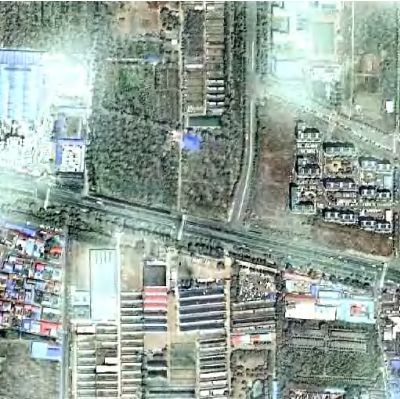}
        \caption*{HL}
    \end{subfigure}
    \begin{subfigure}[b]{0.065\textwidth}
        \includegraphics[width=\textwidth]{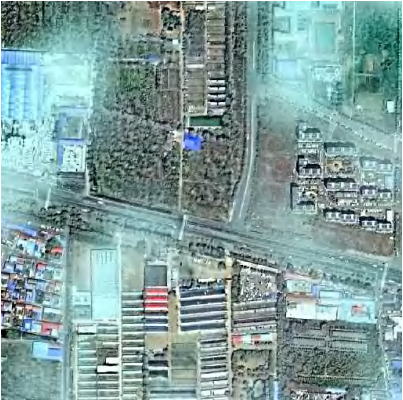}
        \caption*{EVPM}
    \end{subfigure}
    \begin{subfigure}[b]{0.065\textwidth}
        \includegraphics[width=\textwidth]{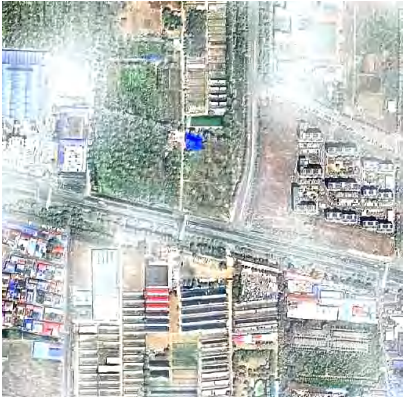}
        \caption*{IDeRs}
    \end{subfigure}
    \begin{subfigure}[b]{0.065\textwidth}
        \includegraphics[width=\textwidth]{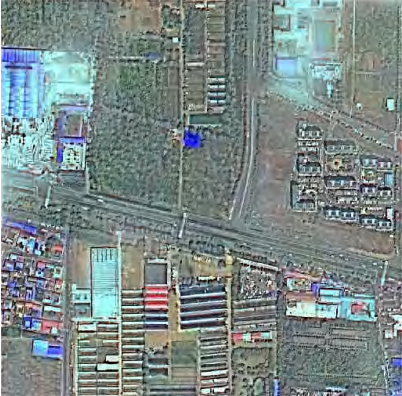}
        \caption*{AOD}
    \end{subfigure}
    \begin{subfigure}[b]{0.065\textwidth}
        \includegraphics[width=\textwidth]{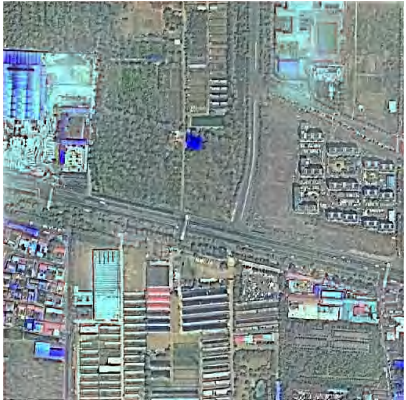}
        \caption*{LDN}
    \end{subfigure}
    \begin{subfigure}[b]{0.065\textwidth}
        \includegraphics[width=\textwidth]{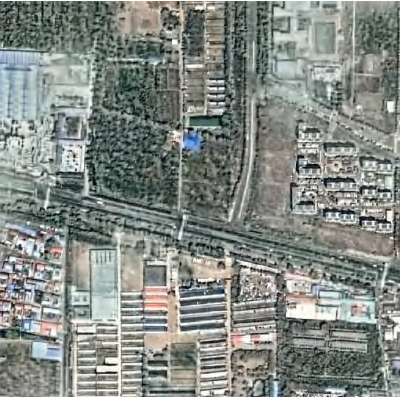}
        \caption*{RSH}
    \end{subfigure}
    \begin{subfigure}[b]{0.065\textwidth}
        \includegraphics[width=\textwidth]{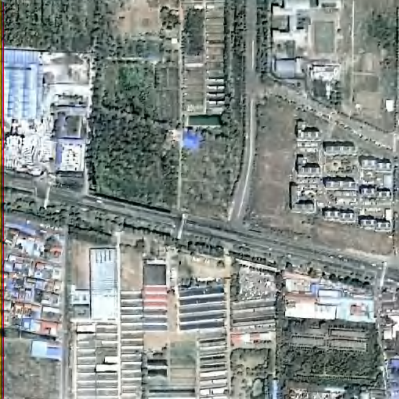}
        \caption*{AUNet}
    \end{subfigure}
    \begin{subfigure}[b]{0.065\textwidth}
        \includegraphics[width=\textwidth]{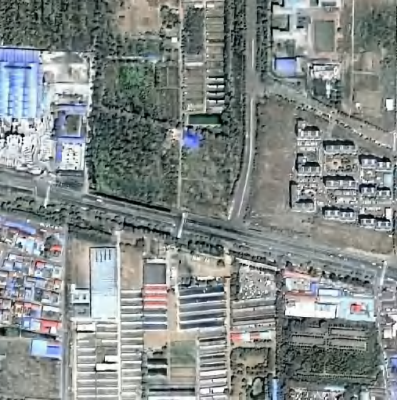}
        \caption*{LRSDN}
    \end{subfigure}
    \begin{subfigure}[b]{0.065\textwidth}
        \includegraphics[width=\textwidth]{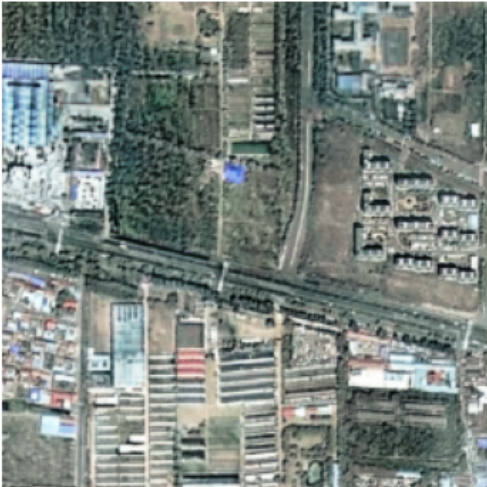}
        \caption*{\textbf{Ours}}
    \end{subfigure}
    \begin{subfigure}[b]{0.065\textwidth}
        \includegraphics[width=\textwidth]{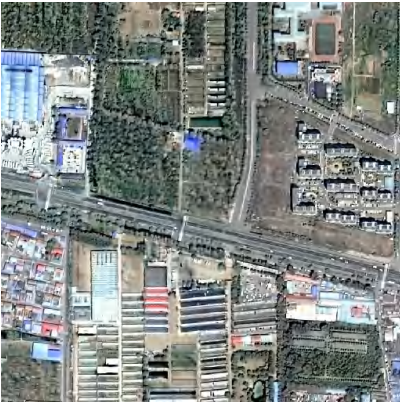}
        \caption*{GT}
    \end{subfigure}
    \label{fig:rice-dehazing-results}
    \caption{Visual comparisons of dehazed results by different methods on Haze1k dataset. (g–i) are
samples from the thin, moderate, and thick haze subsets of the Haze1k test set, respectively}
\end{figure*}
\begin{figure*}[ht]
    \centering
    \begin{subfigure}[b]{0.065\textwidth}
        \includegraphics[width=\textwidth]{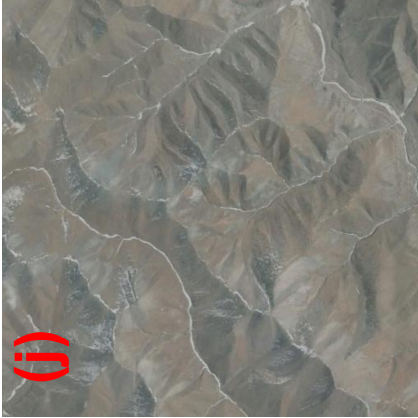}
    \end{subfigure}
    \begin{subfigure}[b]{0.065\textwidth}
        \includegraphics[width=\textwidth]{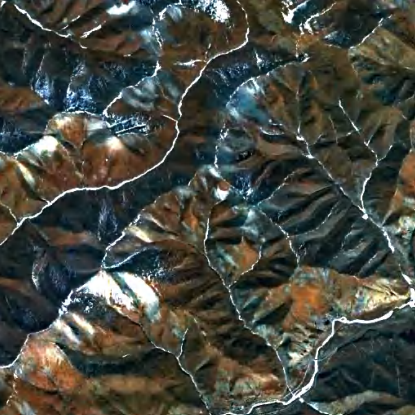}
    \end{subfigure}
    \begin{subfigure}[b]{0.065\textwidth}
        \includegraphics[width=\textwidth]{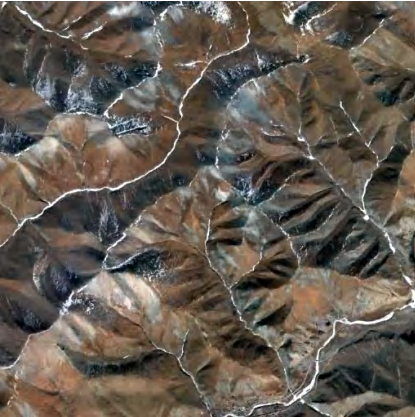}
    \end{subfigure}
    \begin{subfigure}[b]{0.065\textwidth}
        \includegraphics[width=\textwidth]{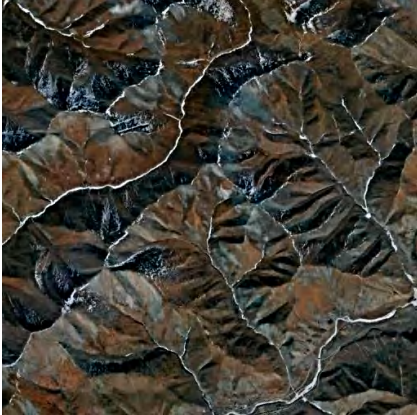}
    \end{subfigure}
    \begin{subfigure}[b]{0.065\textwidth}
        \includegraphics[width=\textwidth]{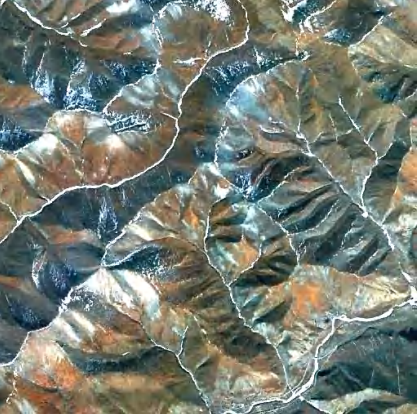}
    \end{subfigure}
    \begin{subfigure}[b]{0.065\textwidth}
        \includegraphics[width=\textwidth]{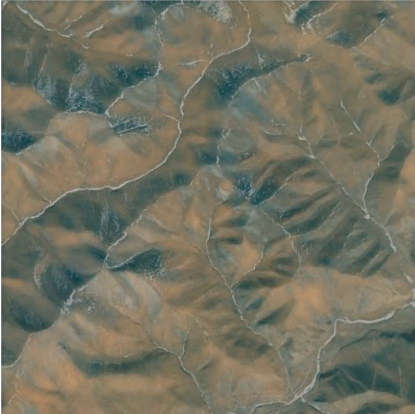}
    \end{subfigure}
    \begin{subfigure}[b]{0.065\textwidth}
        \includegraphics[width=\textwidth]{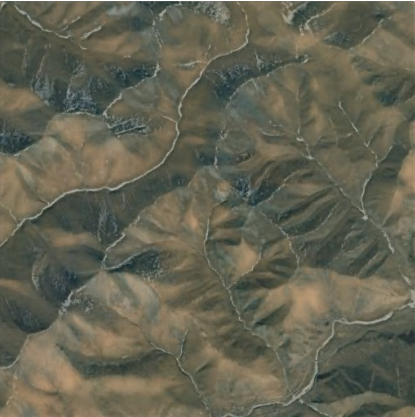}
    \end{subfigure}
    \begin{subfigure}[b]{0.065\textwidth}
        \includegraphics[width=\textwidth]{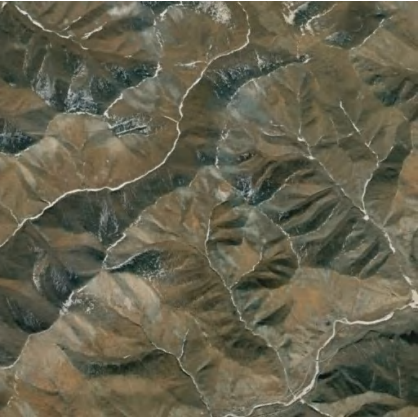}
    \end{subfigure}
    \begin{subfigure}[b]{0.065\textwidth}
        \includegraphics[width=\textwidth]{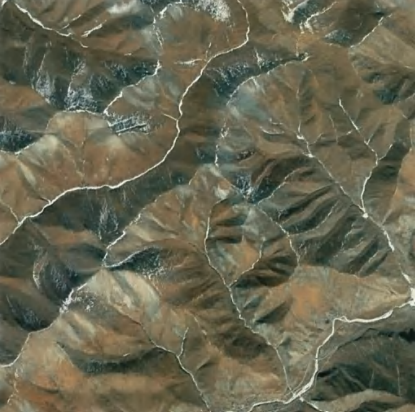}
    \end{subfigure}
    \begin{subfigure}[b]{0.065\textwidth}
        \includegraphics[width=\textwidth]{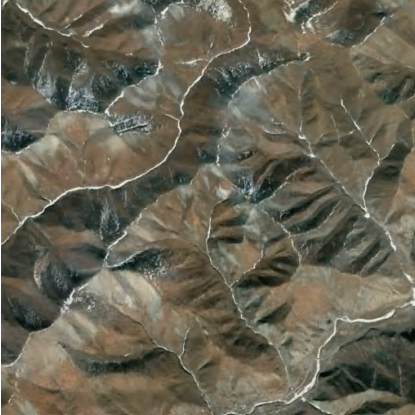}
    \end{subfigure}
    \begin{subfigure}[b]{0.065\textwidth}
        \includegraphics[width=\textwidth]{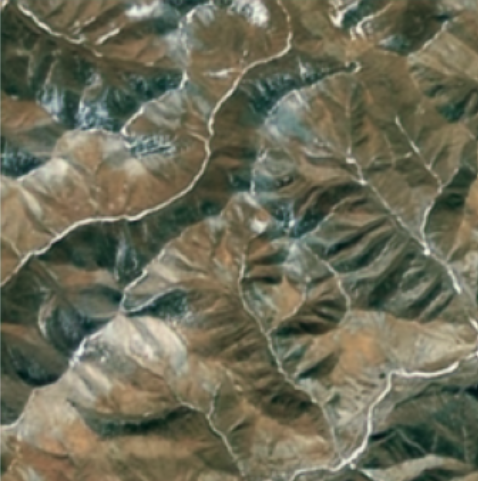}
    \end{subfigure}
    \begin{subfigure}[b]{0.065\textwidth}
        \includegraphics[width=\textwidth]{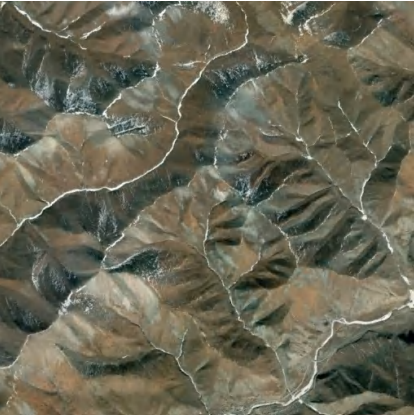}
    \end{subfigure}

    \vspace{0.1cm}

    \begin{subfigure}[b]{0.065\textwidth}
    \includegraphics[width=\textwidth]{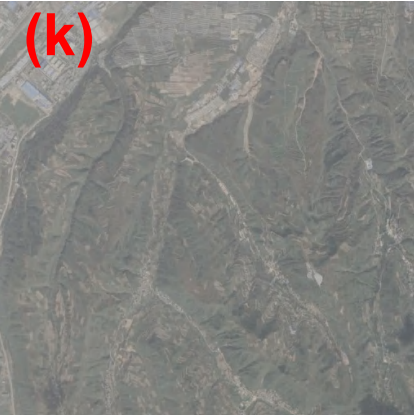}
        \caption*{Input}
    \end{subfigure}
    \begin{subfigure}[b]{0.065\textwidth}
        \includegraphics[width=\textwidth]{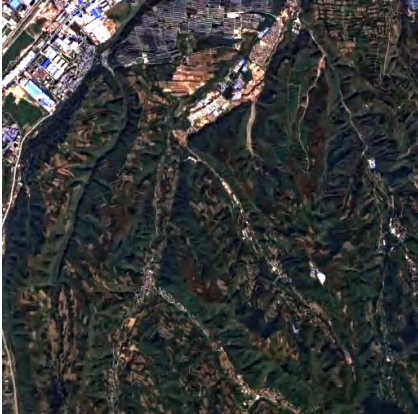}
        \caption*{CEP}
    \end{subfigure}
    \begin{subfigure}[b]{0.065\textwidth}
        \includegraphics[width=\textwidth]{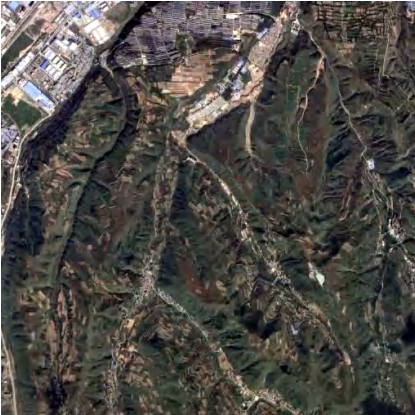}
        \caption*{HL}
    \end{subfigure}
    \begin{subfigure}[b]{0.065\textwidth}
        \includegraphics[width=\textwidth]{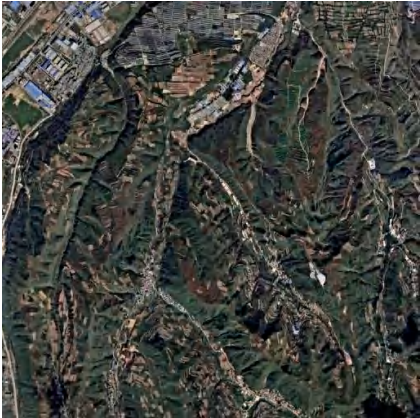}
        \caption*{EVPM}
    \end{subfigure}
    \begin{subfigure}[b]{0.065\textwidth}
        \includegraphics[width=\textwidth]{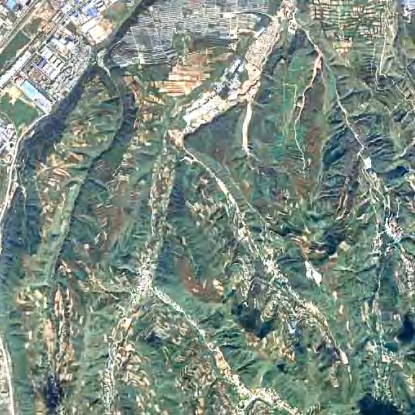}
        \caption*{IDeRs}
    \end{subfigure}
    \begin{subfigure}[b]{0.065\textwidth}
        \includegraphics[width=\textwidth]{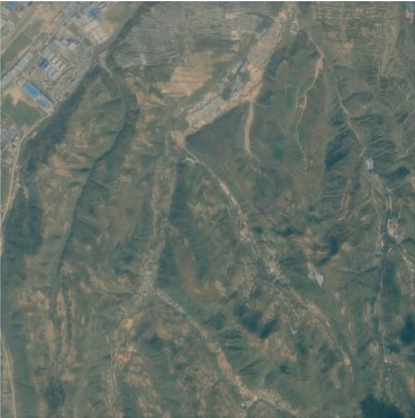}
        \caption*{AOD}
    \end{subfigure}
    \begin{subfigure}[b]{0.065\textwidth}
        \includegraphics[width=\textwidth]{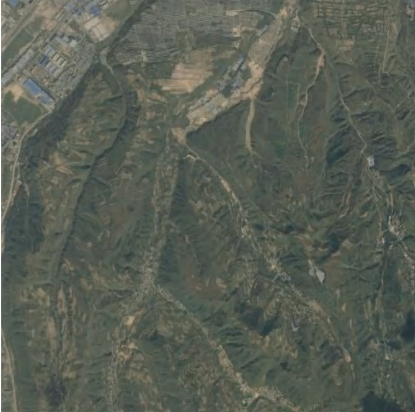}
        \caption*{LDN}
    \end{subfigure}
    \begin{subfigure}[b]{0.065\textwidth}
        \includegraphics[width=\textwidth]{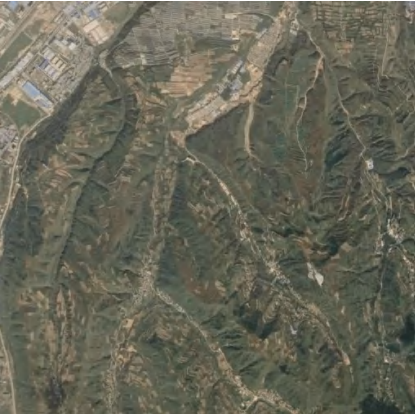}
        \caption*{RSH}
    \end{subfigure}
    \begin{subfigure}[b]{0.065\textwidth}
        \includegraphics[width=\textwidth]{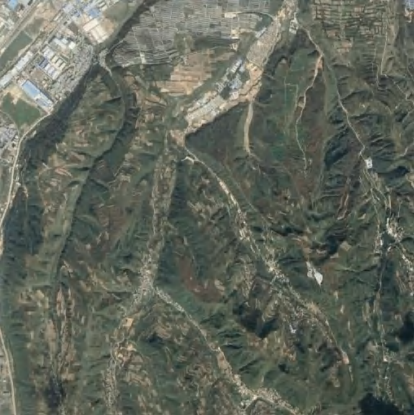}
        \caption*{AUNet}
    \end{subfigure}
    \begin{subfigure}[b]{0.065\textwidth}
        \includegraphics[width=\textwidth]{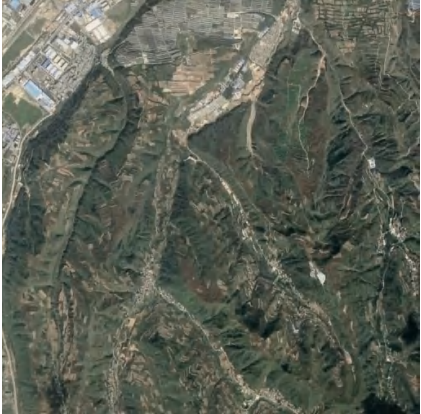}
        \caption*{LRSDN}
    \end{subfigure}
    \begin{subfigure}[b]{0.065\textwidth}
        \includegraphics[width=\textwidth]{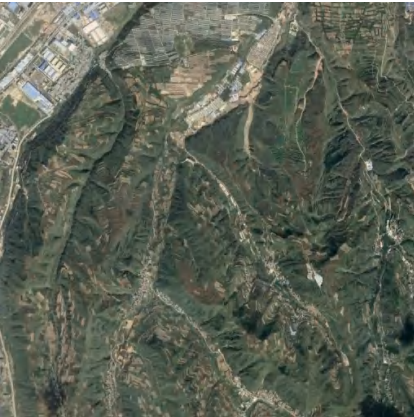}
        \caption*{\textbf{Ours}}
    \end{subfigure}
    \begin{subfigure}[b]{0.065\textwidth}
        \includegraphics[width=\textwidth]{RICE-2/GT.png}
        \caption*{GT}
    \end{subfigure}
    \caption{Visual comparisons of dehazed results by different methods on the RICE dataset.}
\end{figure*}
\subsection{Implementation Details}
IncepDehazeGan is trained on an input hazy image size of $256 \times 256 \times 3$. The network has been trained for $100$ epochs using Adam optimizer for both generator and discriminator. An initial learning rate of $0.0002$ and beta values of $0.5$ and $0.999$  have been used for computing running averages of gradient and its square respectively. IncepDehazeGAN has been trained using  NVIDIA Tesla P100 GPU, with 16GB memory and 3584 CUDA cores.
\begin{table*}[!ht]
    \centering
    \scriptsize
    \caption{Quantitative comparisons of different algorithms' dehazing on the Haze1k dataset.}
    \label{table:dehazing-comparison}
    \begin{tabularx}{\textwidth}{|c|>{\centering\arraybackslash}c|>{\centering\arraybackslash}X|>{\centering\arraybackslash}X|>{\centering\arraybackslash}X|>{\centering\arraybackslash}X|>{\centering\arraybackslash}X|>{\centering\arraybackslash}X|>{\centering\arraybackslash}X|>{\centering\arraybackslash}X|>{\centering\arraybackslash}X|>{\centering\arraybackslash}X|}
        \hline
        \centering
        \textbf{Dataset} & \textbf{Metrics} & \textbf{[7]} & \textbf{[5]} & \textbf{[18]} & \textbf{[19]} & \textbf{[16]} & \textbf{[20]} & \textbf{[15]} & \textbf{[14]} & \textbf{[13]} &\textbf{Ours} \\ \hline
        \multirow{4}{*}{Thick} 
            & LPIPS$\downarrow$ & 0.222 & 0.186 & 0.210 & 0.311 & 0.238 & 0.232 & 0.220 & 0.190 & 0.157 & \textbf{0.088} \\ \cline{2-12}
            & PSNR$\uparrow$ & 15.089 & 16.365 & 16.647 & 11.754 & 16.521 & 16.506 & 20.258 & 19.909 & 21.897 & \textbf{21.612} \\ \cline{2-12}
            & SSIM$\uparrow$ & 0.759 & 0.790 & 0.787 & 0.702 & 0.774 & 0.777 & 0.835 & 0.837 & 0.847 &\textbf{0.937} \\ \cline{2-12}
            & FSIM$\uparrow$ & 0.889 & 0.914 & 0.901 & 0.872 & 0.864 & 0.862 & 0.941 & 0.948 & \textbf{0.959} & 0.956 \\ \hline
        \multirow{4}{*}{Moderate} 
            & LPIPS$\downarrow$ & 0.274 & 0.198 & 0.104 & 0.320 & 0.175 & 0.116 & 0.061 & 0.104 & 0.088 & \textbf{0.055} \\ \cline{2-12}
            & PSNR$\uparrow$ & 13.083 & 15.454 & 20.656 & 14.763 & 20.078 & 20.970 & 24.880 & 24.327 & 25.241 & \textbf{25.731}  \\ \cline{2-12}
            & SSIM$\uparrow$ & 0.746 & 0.798 & 0.918 & 0.785 & 0.906 & 0.921 & 0.941 & 0.929 & 0.934 & \textbf{0.965} \\ \cline{2-12}
            & FSIM$\uparrow$ & 0.854 & 0.895 & 0.942 & 0.899 & 0.917 & 0.936 & 0.966 & 0.956 & 0.970 & \textbf{0.974} \\ \hline
        \multirow{4}{*}{Thin} 
            & LPIPS$\downarrow$ & 0.287 & 0.183 & 0.088 & 0.279 & 0.098 & 0.095 & 0.083 & 0.071 & 0.070 & \textbf{0.044} \\ \cline{2-12}
            & PSNR$\uparrow$ & 12.194 & 13.921 & 20.426 & 15.048 & 18.671 & 18.648 & 22.377 & 23.017 & 23.673 & \textbf{24.445} \\ \cline{2-12}
            & SSIM$\uparrow$ & 0.701 & 0.760 & 0.891 & 0.772 & 0.870 & 0.873 & 0.903 & 0.909 & 0.913 & \textbf{0.965} \\ \cline{2-12}
            & FSIM$\uparrow$ & 0.834 & 0.892 & 0.962 & 0.911 & 0.943 & 0.947 & 0.967 & 0.976 & \textbf{0.978} & 0.975 \\ \hline
    \end{tabularx}
\end{table*}
\begin{table*}[!ht]
    \scriptsize
    \centering
    \caption{Quantitative comparisons of different algorithms' dehazing on the RICE dataset.}
    \label{table:rice-dehazing-comparison}
    \begin{tabularx}{\textwidth}{|c|>{\centering\arraybackslash}X|>{\centering\arraybackslash}X|>{\centering\arraybackslash}X|>{\centering\arraybackslash}X|>{\centering\arraybackslash}X|>{\centering\arraybackslash}X|>{\centering\arraybackslash}X|>{\centering\arraybackslash}X|>{\centering\arraybackslash}X|>{\centering\arraybackslash}X|}
        \hline
        \textbf{Metrics} & \textbf{[7]} & \textbf{[5]} & \textbf{[18]} & \textbf{[19]} & \textbf{[16]} & \textbf{[20]} & \textbf{[15]} & \textbf{[14]} & \textbf{[13]} &\textbf{Ours} \\ \hline
        LPIPS$\downarrow$ & 0.341 & 0.288 & 0.293 & 0.363 & 0.274 & 0.183 & 0.106 & 0.075 & 0.077 & \textbf{0.072}\\ \hline
        PSNR$\uparrow$ & 14.234 & 17.058 & 15.217 & 15.750 & 20.784 & 23.108 & 23.453 & 28.704 & \textbf{31.662} & 29.203\\ \hline
        SSIM$\uparrow$ & 0.713 & 0.723 & 0.742 & 0.611 & 0.834 & 0.873 & 0.919 & 0.946 & 0.953 &\textbf{0.957} \\ \hline
        FSIM$\uparrow$ & 0.800 & 0.781 & 0.865 & 0.746 & 0.856 & 0.887 & 0.961 & 0.980 & \textbf{0.983} & 0.964 \\ \hline
    \end{tabularx}
\end{table*}
\subsection{Performance of Datasets}
\subsubsection{Haze1k}
The Haze1k dataset has 3 sets of testing data with varying haziness, namely thin, moderate and thick, with each set containing 45 images for testing. On this dataset, IncepDehazeGAN achieves an SSIM of 0.965 on the thin and moderate datasets and SSIM of 0.934 on the thick testing dataset.
\subsubsection{RICE}
For training, this dataset was split into a 90:10 ratio with 90\% used for training and 10\% used for testing. Our network achieves an SSIM of 0.9513, PSNR of 28.612 and FSIM of 0.9616.
\subsection{Comparsion with other Models}
Our model delivers remarkable results on both datasets. Starting with Haze1k, where many novel models struggle to achieve high accuracy on the thick test dataset, our model outperformed the current state-of-the-art, LRSDN, by achieving a 10\% higher SSIM. Additionally, we achieved a 43\% lower LPIPS score compared to LRSDN. On the moderate test set, our model reduced the LPIPS score by 37\%. In the RICE dataset, our model achieved results comparable to the state-of-the-art, with only a 2\% variation.
\section{Conclusion}
This paper proposes a novel image dehazing network, IncepDehazeGAN, which utilizes Inception blocks and residual connections to effectively reconstruct clear images from hazy image inputs. Extensive experimental results and comparison with current state-of-the-art models on a variety of metrics (PSNR, SSIM, LPIPS and FSIM) confirm the superiority and effectiveness of our model on the widely used Haze1k and RICE datasets. Grad-CAM XAI technique applied to our network verifies that the model focuses on the hazy regions of the input, as is the intended task of the model.
\newpage
\section{References}
1. Lee, G. Y., Chen, J., Dam, T., Ferdaus, M. M., Poenar, D. P., \& Duong, V. N. (2024). Dehazing Remote Sensing and UAV Imagery: A Review of Deep Learning, Prior-based, and Hybrid Approaches. arXiv [Cs.CV]. Retrieved from \url{http://arxiv.org/abs/2405.07520}
\newline
2. Zhang, S., Zhao, L., Hu, K., Feng, S., En, F., \& Zhao, L. (09 2023). Deep guided transformer dehazing network. Scientific Reports, 13. doi:\url{10.1038/s41598-023-41561-z}
\newline
3. Narasimhan, S. G., \& Nayar, S. K. (2002). Vision and the Atmosphere. International Journal of Computer Vision, 48(3), 233–254.
\newline
doi:\url{10.1023/A:1016328200723}
\newline
4. Narasimhan, S., \& Nayar, S. (12 2015). Interactive (De)weathering of an image using physical models. IEEE Workshop on Color and Photometric Methods in Computer Vision, 10.
\newline
5. Berman, D., Treibitz, T., \& Avidan, S. (11 2018). Single Image Dehazing Using Haze-Lines. IEEE Transactions on Pattern Analysis and Machine Intelligence, PP, 1–1. doi:\url{10.1109/TPAMI.2018.2882478}
\newline
6. Jiang, Y., Sun, C., Zhao, Y., \& Yang, L. (12 2017). Image Dehazing Using Adaptive Bi-Channel Priors on Superpixels. Computer Vision and Image Understanding, 165, 17–32. doi:\url{10.1016/j.cviu.2017.10.014}
\newline
7. Bui, T. M., \& Kim, W. (2018). Single Image Dehazing Using Color Ellipsoid Prior. IEEE Transactions on Image Processing, 27(2), 999–1009. doi:\url{10.1109/TIP.2017.2771158}
\newline
8. Goodfellow, I. J., Pouget-Abadie, J., Mirza, M., Xu, B., Warde-Farley, D., Ozair, S., … Bengio, Y. (2014). Generative Adversarial Networks. arXiv Stat.ML. Retrieved from \url{http://arxiv.org/abs/1406.2661}
\newline
9. Szegedy, C., Liu, W., Jia, Y., Sermanet, P., Reed, S., Anguelov, D., … Rabinovich, A. (2014). Going Deeper with Convolutions. arXiv Cs.CV. Retrieved from \url{http://arxiv.org/abs/1409.4842}
\newline
10. He, K., Zhang, X., Ren, S., \& Sun, J. (2015). Deep Residual Learning for Image Recognition. arXiv Cs.CV. Retrieved from \url{http://arxiv.org/abs/1512.03385}
\newline
11. Xu, G., Wang, X., Wu, X., Leng, X., \& Xu, Y. (2024). Development of Skip Connection in Deep Neural Networks for Computer Vision and Medical Image Analysis: A Survey. arXiv Eess.IV. Retrieved from \url{http://arxiv.org/abs/2405.01725}
\newline
12. Dong, S., \& Chen, Z. (2021). A Multi-Level Feature Fusion Network for Remote Sensing Image Segmentation. Sensors, 21(4). doi:\url{10.3390/s21041267}
\newline
13. He, Y., Li, C., Li, X., \& Bai, T. (2024). A Lightweight CNN Based on Axial Depthwise Convolution and Hybrid Attention for Remote Sensing Image Dehazing. Remote Sensing, 16(15). doi:\url{10.3390/rs16152822}
\newline
14. Du, Y., Li, J., Sheng, Q., Zhu, Y., Wang, B., \& Ling, X. (2024). Dehazing Network: Asymmetric Unet Based on Physical Model. IEEE Transactions on Geoscience and Remote Sensing, 62, 1–12. doi:\url{10.1109/TGRS.2024.3359217}
\newline
15. Wen, Y., Gao, T., Li, Z., Zhang, J., \& Chen, T. (04 2024). Encoder-Minimal and Decoder-Minimal Framework for Remote Sensing Image Dehazing. 36–40. doi:\url{10.1109/ICASSP48485.2024.10446125}
\newline
16. Li, B., Peng, X., Wang, Z., Xu, J., \& Feng, D. (2017). AOD-Net: All-in-One Dehazing Network. 2017 IEEE International Conference on Computer Vision (ICCV), 4780–4788. doi:\url{10.1109/ICCV.2017.511}
\newline
17. Li, Y., \& Chen, X. (2021). A Coarse-to-Fine Two-Stage Attentive Network for Haze Removal of Remote Sensing Images. IEEE Geoscience and Remote Sensing Letters, 18(10), 1751–1755. doi:\url{10.1109/LGRS.2020.3006533}
\newline
18. Han, J., Zhang, S., Fan, N., \& Ye, Z. (2022). Local patchwise minimal and maximal values prior for single optical remote sensing image dehazing. Information Sciences, 606, 173–193. doi:\url{10.1016/j.ins.2022.05.033}
\newline
19. Xu, L., Tree, P., Yan, Y., Kwong, S., Chen, J., \& Duan, L.-Y. (03 2019). IDeRS: Iterative Dehazing Method for Single Remote Sensing Image. Information Sciences, 489. doi:\url{10.1016/j.ins.2019.02.058}
\newline
20. Ullah, H., Muhammad, K., Irfan, M., Anwar, S., Sajjad, M., Imran, A., \& Albuquerque, V. H. C. (10 2021). Light-DehazeNet: A Novel Lightweight CNN Architecture for Single Image Dehazing. IEEE Transactions on Image Processing, PP, 1–1. doi:\url{10.1109/TIP.2021.3116790}
\newline
21. Selvaraju, R. R., Cogswell, M., Das, A., Vedantam, R., Parikh, D., \& Batra, D. (2019). Grad-CAM: Visual Explanations from Deep Networks via Gradient-Based Localization. International Journal of Computer Vision, 128(2), 336–359. doi:\url{10.1007/s11263-019-01228-7}
\newline
22. Huang, B., Li, Z., Yang, C., Sun, F., \& Song, Y. (03 2020). Single Satellite Optical Imagery Dehazing using SAR Image Prior Based on Conditional Generative Adversarial Networks. 1795–1802. doi:\url{10.1109/WACV45572.2020.9093471}
\newline
23. Lin, D., Xu, G., Wang, X., Wang, Y., Sun, X., \& Fu, K. (2019). A Remote Sensing Image Dataset for Cloud Removal. arXiv Cs.CV. Retrieved from \url{http://arxiv.org/abs/1901.00600}
\newline
24. Isola, P., Zhu, J.-Y., Zhou, T., \& Efros, A. A. (2018). Image-to-Image Translation with Conditional Adversarial Networks.
\end{document}